\documentclass[letterpaper, 10 pt, conference]{ieeeconf}  

\IEEEoverridecommandlockouts                              

\usepackage{graphicx} 
\usepackage{amsmath} 
\usepackage{amssymb}  
\usepackage{wrapfig}

\title{\LARGE \bf
A Hybrid Approach for Learning to Shift and Grasp with Elaborate Motion Primitives
}

\author{Zohar Feldman$^{1}$, Hanna Ziesche$^{1}$, Ngo Anh Vien$^{1}$, Dotan Di Castro$^{1}$
\thanks{}
\thanks{$^{1}$ Bosch Center for Artificial Intelligence (BCAI)}
}

\begin{document}

\maketitle
\thispagestyle{empty}
\pagestyle{empty}


\begin{abstract}
Many possible fields of application of robots in real world settings hinge on the ability of robots to grasp objects. As a result, robot grasping has been an active field of research for many years. With our publication we contribute to the endeavor of enabling robots to grasp, with a particular focus on bin picking applications.  
Bin picking is especially challenging due to the often cluttered and unstructured arrangement of objects and the often limited graspability of objects by simple top down grasps. To tackle these challenges, we propose a fully self-supervised reinforcement learning approach based on a hybrid discrete-continuous adaptation of soft actor-critic (SAC). 
We employ parametrized motion primitives for pushing and grasping movements in order to enable a flexibly adaptable behavior to the difficult setups we consider. Furthermore, we use data augmentation to increase sample efficiency. We demonstrate our proposed method on challenging picking scenarios in which planar grasp learning or action discretization methods would face a lot of difficulties. 
\end{abstract}

\section{Introduction}

The most fundamental atomic task of robotic manipulation is grasping. When successful, it enables the robot to do more complex manipulation tasks such as pick and place or bin picking \cite{lenz2015deep}. In the bin picking task, several objects are placed inside a bin, and the goal is to remove all the objects from the bin and place them at a target position.
Bin picking has enormous implications in many areas of robotic manipulation, e.g., industrial assembly, logistics automation, domestic robotics, pick and place tasks, and more. In addition, this task is among one of the most complex robotic manipulation problems, since it exhibits many challenges, yet to be addressed, such as noise and occlusions in perception, object obstructions and collision in motion planning. Therefore, a resilient and robust approach to pick an object out of a bin is needed.

There are two approaches to robotic manipulation. The classical approach is based on planning  \cite{maynard2001maynard,chryssolouris2013manufacturing}. In this approach, a domain expert carefully plans what the robot should do in order to achieve the task at hand. The second, more recent approach consists in learning such skills using machine learning techniques. The planning in this approach is reduced to the necessary requirements and the system learns what to do by adopting trial and error methods such as reinforcement learning  \cite{sutton2018reinforcement}. At one extreme side of the automation-learning spectrum, many recent approaches attempt to solve these tasks using an end-to-end learning approach based on visual input (i.e., "from pixels"; e.g., \cite{levine2016end,zhang2018deep}). 
There are however also methods, that lie somewhere between purely planning and purely learning based approaches. One of these methods is learning based on \emph{motion primitives}. These methods define a set of required primitives to solve the problem at hand \cite{zeng2018learning} where for each primitive there are parameters defining how this primitive is executed. 

\begin{figure}
\centering
\includegraphics[width=.4\textwidth]{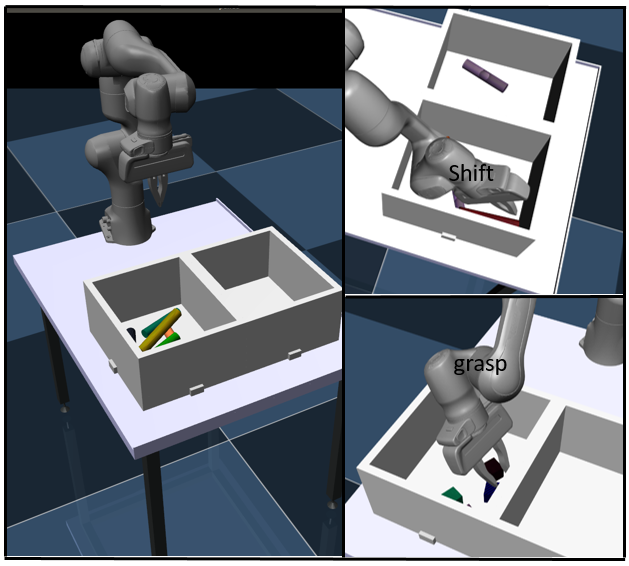}
\caption{\label{fig:setting} The robot is tasked with the goal of emptying one bin to another, shifting the objects (top-right) when necessary in order to facilitate the grasping (bottom-right). The inferred motion primitives have high flexibility to be able to cope in challenging conditions}
\end{figure}
In this work we propose a method based on continuously parametrized motion primitives.  The parameters defining these motion primitives are provided as output of a deep neural network architecture and can be learned in an end-to-end fashion. The advantages for this approach are twofold. \emph{Firstly}, we do not need to apply discretization to the continuous action of the robot. This enables computation and memory efficiency during training, and generalization between similar situations. \emph{Secondly}, our approach allows us to add more parameters to each primitive while avoiding the curse of dimensionality of discretization. Thus we are able to work with actions of a higher degree of freedom. In other words, 
in our approach the output of the network scales linearly in the dimensionality of the actions, while in discretized settings, the size of the output space increases exponentially. 

Specifically, in this work we focus on two primitives: shifting and grasping as depicted in Fig. \ref{fig:setting}. For grasping we propose a parameterization with two discrete parameters (2D position in x-y plane in a heightmap image) and three continuous parameters (yaw, pitch, and gripper width) while for shifting we propose a parameterization with two discrete parameters (2D position in x-y plane in a heightmap image) five parameters (yaw, pitch, roll, shifting direction and shifting distance).
While our algorithm now involves both discrete and continuous actions, we do not need to employ a joint hybrid formulation as proposed in \cite{neunert2020continuous}. Due to the fact that our continuous actions depend on the choice of a discrete action, we can instead resort to a hierarchical reinforcement learning (RL) formulation and policy optimization. Our formulation shares some similarity with recent work \cite{delalleau2019discrete} in RL. However these methods have not been extended to robotic applications, which possess more modeling challenges, e.g. high-dimensional input spaces.

Our main contributions in this work are as follows:
\begin{enumerate}
    \item We present an  algorithm based on Soft Actor Critic (SAC \cite{haarnoja2018soft}) for learning complex grasping and shifting policies directly from pixels.
    \item We devise a bin picking methodology featuring low sample complexity to overcome the physical limitations for successful grasping. This is specifically relevant, when using parallel-jaw grippers on objects in hard to grasp locations.
    \item We propose a novel network architecture, which is designed to model hybrid action spaces, i.e. discrete-continuous actions. 
\end{enumerate}

The paper is organized as follows. In Section \ref{sec:related} we review related work for this paper. In Section \ref{sec:background} we describe the mathematical framework that we use throughout the paper. In Section \ref{sec:method} we describe the methodology used whereas in Section \ref{sec:experimental} we show  the applicability of our work in simulation. We conclude in Section \ref{sec:discussion} and point out future work directions.




\section{Related work}
\label{sec:related}

As a prerequisite for many robot manipulation tasks, grasping objects has been a long-standing subject of interest in robotic research and a large body of literature exists on the topic. While early approaches were mainly based on analytic methods, performing constrained optimization on certain criteria defining stable grasp (see \cite{Bicchi2000RoboticGA} for a comprehensive review), data-driven approaches dominate the field since at least around the year 2000. The survey by Bogh et al. \cite{6672028} provides an excellent overview of these 'classical' data-driven approaches. With the success of deep neural networks, in particular in the domain of computer vision, also the robotic grasping community has more and more turned towards deep learning methods. A substantial amount of publications from the recent years follow the paradigm of supervised learning and cast the problem of robot grasping as one of predicting oriented rectangles from visual inputs. As noted first by \cite{DBLP:journals/corr/abs-1301-3592}, given a depth image five degrees of freedom (DoF) are sufficient to specify a rectangle that fully defines a 7 DoF robot grasping pose (3D position, a 3D rotation, gripper opening width). 


{\bf Grasping via supervised learning}: Distinct approaches differ in how they tackle the huge search space of possible rectangles. Lenz et al. \cite{DBLP:journals/corr/abs-1301-3592} suggest a two step approach relying on one fast but inaccurate step, performing an exhaustive search over the space of possible rectangles, followed by a slower accurate step to re-rank the top 100 rectangles. In contrast, Redmon et. al. \cite{redmon2015realtime} and Kumra et al. \cite{kumra2017robotic} propose a faster one-step approach, superseding the exhaustive search step. More recent approaches \cite{doi:10.1177/0278364919859066, DBLP:journals/corr/abs-1909-04810} regress rectangles in a generative fashion by predicting pixel-based grasp angles, gripper width and grasp success scores and using these images as input to a final network predicting the grasp rectangles. Instead of using manually labeled images like in those work, Mahler et al. \cite{mahler2017dex} combined a grasp sampling heuristics with an analytic method to determine force closure based on wrench space analysis in order to generate a large data base of object meshes annotated with grasp poses and corresponding grasp quality metrics \cite{mahler2016dex}. A similar approach was proposed by Schmidt et al. \cite{8463204}. More recent approaches \cite{mousavian20196,cvpr/FangWGL20,sundermeyer2021contact} propose a grasp sampling network that is based on a variational autoencoder, and trained with simulated or annotated data.

{\bf Grasping via self-supervised learning}: Another solution to the problem of training data is provided by direct self-supervision, i.e. by running a prescribed grasping policy either in simulation or on a real robot to collect training data annotated automatically by the observed grasp success. This method was pioneered on a large scale by Pinto et al. \cite{DBLP:journals/corr/PintoG15}. Levine et al. \cite{DBLP:journals/corr/LevinePKQ16} combined this strategy with a vision-based servoing-mechanism to guide the robot to successful grasps and scaled the approach further to a multi-robot distributed setup. In a subsequent publication, the approach was instead formulated as a RL problem \cite{DBLP:journals/corr/abs-1806-10293}. Khansari et al. \cite{khansari2020action} built upon this work and extended it with an image-based action representation, which they found to increase sample efficiency. Furthermore, they trained purely in simulation and used data augmentation to successfully transfer to the real world. In a recent work, Song et al. \cite{song2020grasping} also propose an extension of \cite{DBLP:journals/corr/abs-1803-09956} to full 6D pose grasping. In contrast they however do no longer consider non-prehensile manipulations. Furthermore they rely on a manually collected grasping dataset, while our approach is trained in a fully self-supervised manner. 

{\bf Grasping via non-prehensile manipulation}: Although some of the methods mentioned above have been demonstrated or even explicitly trained in cluttered scenes, grasping actions may not always be enough in the presence of multiple objects, occlusions or workspace limitations. Therefore, another stream of research has focused on combining grasping with non-prehensile actions. Dogar et al. \cite{Dogar-2012-7507} propose a fully planning-based approach using non-prehensile motion primitives to extract an object from clutter. Boularis et al. \cite{10.5555/2887007.2887192} formulate a finite dimensional Markov decision process (MDP) and use tabular Q-learning to train a policy combining pushing and grasping. Their approach relies on hand-crafted features and kernel methods to model the value function, reward and transition function. 
In \cite{DBLP:journals/corr/PintoG16} the authors investigate multi-task supervised learning to perform grasping, pushing and poking actions and find that the multi-task model shows better performance than single-task models trained on the same amount of data. 
A hierarchical approach to combine pushing and grasping primitives has been considered by Danielczuk et al. \cite{DBLP:journals/corr/abs-1903-01588}, who propose a high-level planner that queries the low-level motion primitive-based policies and chooses actions based on the returned quality estimates. In a subsequent publication \cite{DBLP:journals/corr/abs-1903-01588} the authors train a visuomotor policy to uncover objects in clutter in order to increase graspability. To facilitate the difficult exploration problem, the authors define several teacher heuristics. Most related to our work are the approaches proposed by Zeng et al. \cite{DBLP:journals/corr/abs-1803-09956} and by Berscheid et al. \cite{DBLP:journals/corr/abs-1907-11035}. Both define motion primitives for pushing and grasping actions and learn them using Q-learning on a discretized action space consisting of a 2D Cartesian grasping position in the (x,y)-image plane and a yaw rotation around the z-axis. Both methods, however, are restricted to top-down grasps with discrete yaw rotations. A recent work from Berscheid et al. \cite{bs-2103-12810} proposes a hybrid approach of learning  6 DoF grasping. However this method is still simplified to planar grasping in which lateral DoF ($z$-depth, roll, and pitch angles) are inferred from a model-based controller. In our work, we built on these approaches but extend them to allow for full end-to-end 6DoF grasp learning and more flexible motion primitives. We achieve this by giving up the strict discretization in favour of a hybrid discrete-continuous action space.

\section{Background}
\label{sec:background}
Recent years have shown great improvements in model-free deep reinforcement learning in several domains. In particular SAC (Soft Actor-Critic) \cite{haarnoja2018soft} has gained popularity for continuous control settings due to its sample efficiency and stability and its superior performance both in simulation benchmarks and in real world settings \cite{DBLP:journals/corr/abs-1812-05905}.  For these reasons and due to the fact that SAC also has shown good performance when trained on pixel inputs \cite{kostrikov2020image}, we use SAC as backbone algorithm for our approach. 
SAC is an off-policy actor-critic algorithm and as such jointly trains a pair of state-action value functions $Q^{\pi}_{\phi_i}$, $i=1,2$ and an stochastic policy $\pi_\theta$. As SAC is based on the paradigm of maximum-entropy RL, the actor is trained to maximize the cumulative expected return while at the same time maximizing its entropy, i.e. acting as stochastic as possible. 
In standard SAC, the actor is parametrized as a Gaussian policy $\pi_\theta$ and is trained on the following loss function: 
\begin{align*}
    \mathcal{L}(\theta) = &\mathbb{E}_{a\sim\pi_\theta}\left[Q^\pi(s, a) - \alpha \log \pi_\theta(a|s)\right],
\end{align*}
where $Q^{\pi}(s, a) = \underset{i=1, 2}{\mathrm{min}}Q_{\phi_i}^{\pi}(s, a)$, and $\alpha$ is a coefficient defining the weight between entropy and reward. 
The critics $Q_{\phi_i}$ are trained via Double Deep Q-learning \cite{FujimotoHM18} with targets provided by corresponding temporally delayed target networks $Q_{\bar{\phi}_i}$, i.e the critic loss is given by
\begin{align*}
    \mathcal{L}(\phi_i) = \mathbb{E}_{\overset{s, a, s', r\sim \mathcal{D}}{ a'\sim \pi_\theta(\cdot|s')}}\left[\left(Q_{\phi_i}(s, a) - \left(r + \gamma \,y_t(s', a')\right)\right)^2\right]
\end{align*}
where $y_t(s', a')=\underset{i=1, 2}{\mathrm{min}}Q_{\bar{\phi}_i}(s', a') - \alpha \log \pi_\theta(a'|s')$. 
\newline
Here, states $s$, actions $a$, next states $s'$ and rewards are sampled from a replay buffer which is continually populated as the training progresses. The actions $a'$ in state $s'$ are sampled from  the current policy. 
In their subsequent publication \cite{DBLP:journals/corr/abs-1812-05905} Haarnoja et al.  also propose a method to automatically tune the hyperparameter $\alpha$, which implicityly controls the the policy's exploration.

\section{Method}\label{sec:method}
We model the Bin Picking task as a finite-horizon Markov Decision Process (MDP;  \cite{puterman1994markov}) $\left(\mathcal{S}, \mathcal{A}, \mathcal{T}, r, \gamma, H\right)$ with state space $\mathcal{S}$, action space $\mathcal{A}$, transition probability function $\mathcal{T}$, reward function $r$, discount factor $\gamma$, and horizon of $H$ steps. In each step $t=1,\ldots,H$, a state $s_t\in\mathcal{S}$ is observed, and an action $a_t\in\mathcal{A}$ is chosen according to a policy $\pi(a_t|s_t)$. Upon the application of $a_t$ in $s_t$, a reward $r\left(s_t,a_t\right)$ is received, and the system transitions into a new state $s_{t+1}$ according to $\mathcal{T}$. 
We represent the \textbf{state} $s_t$ by a heightmap image with four channels, the color (RGB) and height (Z) from the table on which the robot and bin are installed. The heightmap is computed from an image captured by RGB-D camera overlooking the bin area. Given the camera intrinsic and extrinsic parameters, the image is transformed to a color point-cloud in the robot coordinate system, whose origin is conveniently located at the base link attached to the table and z-axis points in the inverse direction of gravity. The point-cloud is then projected orthographically onto a 2-dimensional grid with granularity of $5\times5$mm in the xy-plane containing the bin. The \textbf{action} $a_t$ is a motion primitive corresponding to either a grasp or a shift, each defined by a distinct set of parameters, as detailed in Section \ref{ssec:primitives}. A \textbf{reward} $r_t$ of $1$ is received in step $t$ if $a_t$ results in a successful grasp, otherwise it is $0$.
We use RL to train the policy $\pi\left(a_t|s_t\right)$ to maximize the $Q$-function defined as
\begin{equation*}
    Q(s_t, a_t) \triangleq \mathbb{E}_{s'\sim \mathcal T(s'|s,a)}\left[\sum_{i=t}^{H} \gamma^i r(s_i,a_i)\right].
\end{equation*}
The Bellman equation, 
\begin{equation}\label{eq:bellman}
    Q_t(s_t, a_t) = \mathbb{E}_{\pi,\mathcal T} \left[r(s_t,a_t) +  Q_{t+1}(s_{t+1}, a_{t+1})\right],
\end{equation}
provides a means to evaluate the policy $\pi$ by calculating the $Q$-function recursively, and lies in the core logic of most (off-policy) RL algorithms, including the algorithm presented here. 

\subsection{Motion Primitives}\label{ssec:primitives}
In each step the policy outputs a primitive type $\phi\in\left\{\textbf{g}\text{(rasp)}, \textbf{s}\text{(hift)}\right\}$ as well as the primitive parameters sets $a^{\phi}$ that together define the maneuver to be executed. Each primitive is executed essentially in open-loop as follows.
\newline
\textbf{Grasp:} The gripper is first oriented according to the inferred Euler angles $\left(i^g,j^g,k^g\right)$. Subsequently, the gripper tool-center-point (TCP) is moved to a point above the target position defined by the Cartesian coordinates $\left(x^g,y^g,z^g\right)$ and then lowered to the target height $z^g$ having the width between the fingers set to $w^g$ cm. Upon reaching the target pose or detecting a collision, the gripper is closed and lifted up $20$cm at which point the gripper is again signaled to close. The grasp is considered successful if the distance read between the fingers exceeds a threshold set to a value slightly below the smallest size of all objects in consideration. The parameter set $a^{\text{g}}=\left(x^g,y^g,j^g,k^g,w^g\right)$ includes all the above parameters, except $z^g$ which is extracted from the heightmap in position $\left(x^g,y^g\right)$, and the roll angle $i^g$ which is set to $0$ to ensure equal finger heights for stable top-down grasping.
\newline
\textbf{Shift:} The gripper TCP is moved to a target pose $\left(x^s,y^s,z^s,i^s,j^s,k^s\right)$ with clamped fingers, after which it is moved $\overrightarrow{d^s}$ cm in a horizontal direction defined by a rotation angle $\overrightarrow{k^s}$ around the z-axis. Here, $a^{\text{s}}=\left(x^s,y^s,i^s,j^s,k^s,\overrightarrow{d^s},\overrightarrow{k^s}\right)$ whereas, similarly to the grasp primitive, $z^s$ is extracted from the heightmap.

\subsection{Reinforcement Learning for Bin Picking}\label{ssec:algorithm}
\subsubsection{Architecture}
\begin{figure*}[!ht]
\centering
\includegraphics[width=.6\textwidth]{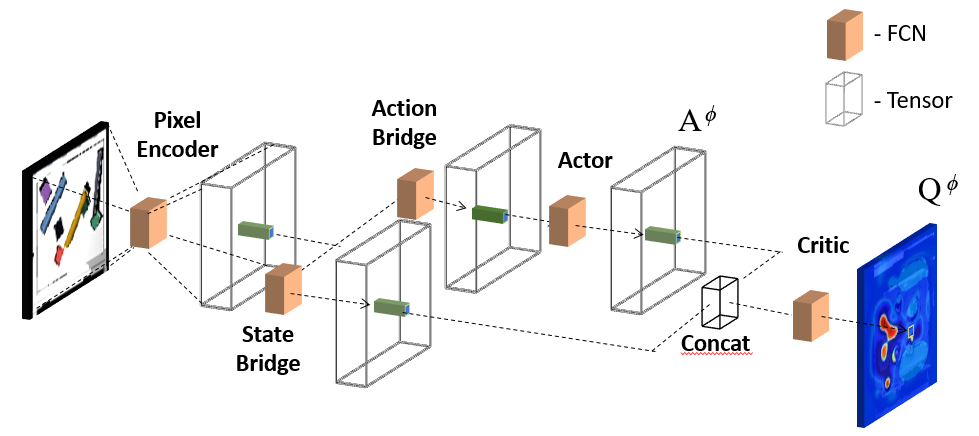}
\caption{\label{fig:Sys} Action selection network: Both grasping and shifting primitives $\phi$ share this architecture. The Pixel Encoder FCN module encodes the spatial features of each of the image pixels, whereas all other FCN modules act as Fully Connected Layers in a context of a single pixel using kernels of size $1$.}
\end{figure*}

We use a Fully Convolutional Network (FCN) to infer the parameter set $a^{\phi}$ and approximate the value $Q^{\phi}(s, a^{\phi})$ for each primitive type $\phi$ and heightmap image $s$. The underlying algorithmic framework and network architecture follows what can be schematically viewed as a combination of convolutional SAC for continuous actions and $Q$-learning for discrete actions. Namely, the network infers an embedding for each of the heightmap pixels using a \textbf{Pixel Encoder} module. By viewing each pixel embedding as a distinct state, an \textbf{Actor} module convolutes over these "states" and infers a Gaussian action for each of them, resulting in an action map $A^{\phi}$. Actions are concatenated with their corresponding pixel embedding and evaluated by a \textbf{Critic} module, resulting in a $Q$-value map $Q^{\phi}$. Figure \ref{fig:Sys} depicts the network architecture which is shared between both primitives. 
Notably, each pixel $(h',w')$ in the action map $A^{\phi}$ induces a fully specified parameters set $a^{\phi}$. The spatial parameters $(x^{\phi},y^{\phi})$ are derived from the location association of the heightmap pixels, whereas the rest of the parameters are given by $A^{\phi}[h',w',:]$, the values in the action map corresponding to the pixel coordinates along all channels. Similarly, $Q^{\phi}[h',w',1]$ represents the $Q$-value of the state-action pair $(s,a^{\phi})$. The  map $Q^{\phi}$ thus represents a $Q$-function $Q^{\phi}(s,a^{\phi})$ for a discrete set of actions corresponding to the heightmap pixels, and as such, it is trained using a $Q$-learning schema for discrete actions. The actor, on the other hand, is trained using a soft actor-critic schema. Figure \ref{fig:Sys} depicts this action selection process. Note that for training the Q-network, the Actor module needs not be executed. Instead, we can concatenate action parameters $a^\phi$ stored in a replay buffer with the pixel embedding of the corresponding state $s$ and then evaluate it using the critic to predict a $Q$-value.

\subsubsection{Inference} Inferring an action for a given state $s$ entails computing all the network modules end-to-end, producing the action map $A^{\phi}$ and $Q^{\phi}$-map for both primitives. The primitive type is selected as $\phi^{\ast}=\arg\max_{\phi}\max_{h',w'}Q^{\phi}[h',w',1]$. The primitive parameters set is calculated by first selecting the best pixel for $\phi^{\ast}$, $(h^{\ast},w^{\ast})=\arg\max_{h',w'}Q^{\phi^{\ast}}[h',w',1]$ and extracting $a^{\phi^{\ast}}$ from $(h^{\ast},w^{\ast})$ as described above. 

\subsubsection{Training} We collect data throughout the experiments and store it in a replay memory. When sampling mini-batches for training, we use \textbf{data augmentation} in order to increase the sample efficiency \cite{kostrikov2020image}. Particularly, we create task-invariant versions of the sampled experiences $\left(s_t,a_t,r_t,s_{t+1}\right)$ by rotating the heightmap image $s_t$ by a random angle, as well as rotating the relevant angles in $a_t$ by the same angle. In both primitives, we rotate the yaw angle $k^{\phi}$, while in the shift primitive we also rotate the shift direction $\overrightarrow{k^s}$. By that, we create new samples that induce similar results, that is, similar $r_t$ and $s_{t+1}$. 
We train each of our models using the following loss functions:
\newline
\textbf{Critic Loss}:
\vspace{-2mm}
\begin{equation*}
    \mathcal{L}_{\text{critic}} = \begin{cases}                                    \text{BCE}\left(Q_t^{\phi}(s_t,a_t^{\phi}), y_t\right)& t=H\\
                                    \text{MSE}\left(Q_t^{\phi}(s_t,a_t^{\phi}), y_t\right)& \text{otherwise}
                                  \end{cases},
\end{equation*}
where $y_t=r_t+\gamma\max_{\phi,a}Q_{t+1}^{\phi}\left(s_t,a\right)$, BCE is the Binary Cross Entropy loss, and MSE is the Mean Squared Error loss.
$H$ denotes a fixed predefined horizon. Note that we use the BCE loss for the final step only , since the corresponding ground label  $y_{H} = r_{H}\in \{0, 1\}$. For all other steps $t<H$, $y_t$ is a real number and we thus use the MSE loss.
The critic loss is minimized with respect to the network parameters of the Pixel Encoder, Pixel Action Encoder, and Critic modules.
\newline
\textbf{Actor Loss}:
\vspace{-2mm}
\begin{equation*}
    \mathcal{L}_{\text{actor}} = Q_t^{\phi}(s_t,a_t^{\phi}) - \alpha\log\pi_t^{\phi}(a_t^{\phi}|s_t)
\end{equation*}
The actor loss is minimized with respect to the parameters of the Pixel Encoder and Actor modules.

\subsection{Implementation Details}\label{ssec:impl}  
\paragraph{Finite Horizon} In finite-horizon MDPs, the Q-function is time-dependent and, respectively, the Q-functions in the different steps should be approximated by distinct networks to be theoretically sound. This, however, entails training $H$ neural networks, which may  impose a large computational burden. Most RL implementations circumvent this issue by treating the MDP as infinite-horizon, regardless of the actual case, and using the discount factor to moderate the effect of future steps. Here, we choose to use distinct networks for the different steps and instead take different relaxing measures. Regardless of the number of steps allowed to empty the bin, we use a fixed and small horizon of $H=2$ and train two networks, $\Phi_0$ and $\Phi_1$. While this choice helps alleviating the hurdles mentioned above which are even amplified by having sparse rewards and large action space, it can also be motivated by the observation that bin picking typically does not benefit much from looking more than a few steps ahead. Indeed, looking ahead beyond the current state is mostly beneficial when shifting is required to allow a consecutive grasp, and most likely a single shift manoeuvre should suffice in this case. Respectively of this relaxation, we use $\Phi_0$ to infer an action in step $t<H$ and $\Phi_1$ for $t=H$.
During training, we use all the recorded experiences for updating the networks of all steps independently of the actual step they were performed within the episode. 

\paragraph{Exploration Heuristics}
In order to increase the chances of seeing an effective result for either a grasp attempt or a shift attempt when taking exploration steps, we use a change detection procedure as described in \cite{Finman-2013-7775} to locate pixels that correspond to objects. Very roughly, we subtract the point-cloud of the current image state from a reference point-cloud of an image with empty bin and mask pixels in which there is a notable difference. We then sample a portion of the exploration actions from a uniform distribution over all the object pixels. We also assume to have at our disposal a bounding box of the bin, which is either known or can be obtained by some external perception utility. We define sites of  interest  on the gripper and wrist mounted camera, and transform these points according to any candidate target pose to verify its feasibility by checking if the transformed points fall inside the bounding box of the bin. If there is at least one point that falls outside the bin, we know that a reach attempt would result in a collision and therefore discard the attempt in advance. We also use this computation as complementary exploration heuristic for searching feasible orientations for a given translation. 

\paragraph{Mini-Batch Sampling} To overcome the impact of sparse rewards on training, particularly as early stages, we sample mini-batches that are comprised of half success samples and half failures.

\section{Experiments}\label{sec:experimental}
\subsection{Simulation environment}
In order to assess the performance of our algorithm in a binpicking scenario, we set up a suitable MuJoCo \cite{todorov2012mujoco} environment, as depicted in Fig.~\ref{fig:setting}. This environment contains a Franka Emika Robot 
with a Realsense d415 camera mounted on its wrist. Two bins of particularly small size $30\,\text{cm}\times 40\,\text{cm}$ and with a height of 20\,cm  are placed in front of the robot. We purposely create this challenging setting to evaluate the limit of different approaches, and demonstrate the benefit of high DoF grasp learning methods like ours. First, the small size of the bin renders direct grasping infeasible for a large fraction of the area inside the bin. Secondly, the high bin walls preclude top-down grasps for many of the remaining graspable object positions, as these would lead to collisions with the walls. Lastly, the wrist mounted camera adds to the difficulty of avoiding collisions.
We use this setup both for training the algorithm and for evaluation. During all experiments we spawn 4 objects from a predefined set at a random position above one bin at the beginning of each episode.  In each step we execute the action returned from the policy $\pi$ as described in Section \ref{sec:method}. We use ROS for control and execute all motion commands using a Cartesian impedance controller. 

\subsection{Experimental setup}
We conduct several experiments to evaluate our algorithm and compare it to one following baseline, called \emph{Learning shifting for grasping} proposed by Berscheid et al. \cite{berscheid2019robot}. Similar to Zeng et. al. \cite{zeng2018learning}, this baseline resorts to action discretization and is restricted to planar grasps, i.e. allows only for rotations around the axis perpendicular to the table top. With our experiments we aim to answer primarily the following questions: 
\begin{itemize}
	\item Do the \emph{non-prehensile shifting actions} improve the performance of our bin picking algorithm, in particular in challenging settings?
	\item Do our flexible and continuously parametrized motion primitives show an advantage over \emph{fully discrete parametrizations}?
	\item How well can our learned solution \emph{generalize to unseen objects}?
\end{itemize}
In order to answer these questions, we set up two different scenarios. All scenarios share the same general setup described above and  differ only in the way the objects are placed into the bin. In particular, we consider the following two settings:
(1) \textbf{Randomized bin}: We spawn objects at random positions above the bin and let them drop into it. This scenarios is also used for training, and (2) \textbf{Close-to-wall bin}: We spawn objects only at random positions close to the bin walls. This is a very challenging setting as the spawn objects are likely very close to the bin wall, so grasping such objects often requires shifting. In other cases if an object can be grasped without a support of shifting, its grasp pose might be harder to find as it involves more constraints, e.g. stable grasp pose, collision avoidance to the bin walls and nearby objects.

The first setting allows us a general evaluation of our algorithm in a challenging setup and a comparison to the chosen baselines. The second setting is designed to test our algorithm in situations where direct grasps and top-down grasps are impossible and therefore flexible shifting and grasping motions are necessary. For each of the test scenarios and algorithms we execute 100 runs and compare average performance according to the following metrics, proposed by Zeng et al. \cite{zeng2018learning}:
\newline
\textbf{Clearance}: percentage \% of objects cleared from the bin
\newline
\textbf{Completion rate}: percentage \% of runs in which the robot succeeds in emptying the bin without executing 10 successive unsuccessful actions.
\newline
\textbf{Success rate}: average percentage \% of successful grasps per episode.
\newline
\textbf{Action efficiency}: defined as $\frac{\text{\# objects in scenario}}{\text{\# actions before completion}}$ this metric measures the capability of the policy to avoid unnecessary actions.
\newline
Another important metric is the compute time required for inferring the best action from the camera input. Our method takes $\approx 0.01s$ when run on a GeForce RTX 2080. It is thus real time capable and would even allow for a closed-loop setup. 

 
\begin{table}[t]
    \centering
    \caption{Experimental Results for the Random Setting (Mean $\%$)}
    \label{tab:results_random}
    \begin{tabular}{c|c|c|c|c}
        Method & Clearance & Completion  & Grasp Succ.  & Action Eff.\\
        \hline
        \cite{berscheid2019robot}& $72$ & $32$       & $30$ & $70$
        \\
        Ours  & $\textbf{96}$ & $\textbf{85}$       & $\textbf{80}$  & $\textbf{76}$  \\
        \hline
        \cite{berscheid2019robot} (G) & $68.3$ &    $21.7$   & $27.4$ & $64.5$
        \\
        Ours (G)  & $\textbf{85}$ & $\textbf{51}$       & $\textbf{55}$  & $\textbf{85}$  \\
        \hline
        \cite{berscheid2019robot} (U)& $31.9$ &    $1.5$   & $43.5$ & $41.9$
        \\
        Ours (U)  & $\textbf{82.4}$ & $\textbf{53}$       & $\textbf{46}$  & $\textbf{73}$ 
        \\
    \end{tabular}
\end{table}

\begin{table}[t]
    \centering
    \caption{Experimental Results for the Challenging Setting (Mean $\%$)}
    \label{tab:results_challenging}
    \begin{tabular}{c|c|c|c|c}
        Method & Clearance & Completion  & Grasp Succ.  & Action Eff.\\
        \hline
       \cite{berscheid2019robot}  & $48$       & $1$ & $15$ & $\textbf{66}$  
        \\
        Ours &  $\textbf{88}$ & \textbf{66}       & $\textbf{40}$  & $34$  
        \\
        \hline
       \cite{berscheid2019robot} (G) & $23.8$ &    $1.6$   & $7.7$ & $62.3$  
        \\
        Ours (G)& $\textbf{72}$ &    $\textbf{30}$   & $\textbf{33}$ & $\textbf{72}$ 
        \\
        \hline
       \cite{berscheid2019robot} (U) & $33.7$ &    $0.9$   & $11.6$ & $\textbf{80}$
        \\
        Ours (U)  & $\textbf{77}$ & $\textbf{49}$   & $\textbf{39}$  & $62$ 
        \\
        \hline
    \end{tabular}
\end{table}

\subsection{Evaluation results}
Tables \ref{tab:results_random} and \ref{tab:results_challenging} summarize the results of the random and challenging settings, respectively. In the basic experiments, the robot is presented with the same object types that were used for training, cubes and rods. In addition, we test each algorithm in modifications. In the first (G), we allow only grasp actions, while in the second (U), the robot is given unseen objects, specifically cans, elongated boxes, balls, and tubes.
As expected, in the challenging setting the performance was worse than the in the random setting emphasizing the fact that most failure cases result from objects being placed at location from which they cannot be grasped with direct top-down grasping. It is also apparent that the method we propose gave better results than \cite{berscheid2019robot}. Here, the relatively tall bin renders top-down grasping and shifting practically impossible in a large area of the bin, as opposed to the setting described in \cite{berscheid2019robot}. Therefore grasping and shifting attempts of objects near the bin walls were mostly ineffective. We also observed that most of these hard cases were solved by applying 6DoF grasps which allow the gripper to reach objects without colliding with the bin wall, and much less by shifting. The typical failure cases were concerned with either not being able to learn the right shift approach, having repeated attempts in misleading cases (e.g. trying to grasp to close rods that appear as one), and in some less common occasions having spawned the objects outside the bin. It is also apparent that our solution is preferable in respect to generalization. While both algorithms show performance degradation in presence of unseen objects, the impact is significantly lower on our algorithm that the baseline. Notably baseline sometimes shows better action efficiency in the challenging setting. This is due to fact that only completed scenes contribute to this metric and that the baseline is not able to solve most of the scenes that require many shifts and therefore necessarily come with a low action efficiency.

\section{Conclusion}
\label{sec:discussion}

This paper proposes a unified actor-critic framework for learning continuous and high-DoF pushing and grasping policies in robotic bin-picking. Bin-picking on industrial applications requires non-trivial grasping skills, e.g. avoiding collisions with the bin wall and nearby objects. It additionally requires non-prehensible actions, e.g. shifting. Existing methods often resort to a discretization of the action space. As a result, they scale poorly to high-dimensional action spaces and cannot easily be extended to 6DoF grasp learning or even higher DoF shifting skill learning. The main contributions of our proposed framework can be summarized as follows. \emph{First}, this framework easily allows more DoFs for the action space based on a well-known actor-critic RL framework. Hence it enables continuous actions efficiently. \emph{Second}, we made an important contribution on the choice of the network architecture with which we can easily model hybrid action spaces, i.e. discrete-continuous actions. The discrete actions are primitives (grasping, shifting), and grasp/shift centers ($xy$-planar), while the continuous actions are grasp/shift configurations.

We demonstrate the proposed framework on a challenging setting which is common in many industrial applications, e.g. featuring a small bin, a high wall bin, ubiquitous collisions etc. This setting requires both high DoF grasp/shift skill configurations, and non-prehensible shifting actions. We showed that our approach is able to complete the tasks with  higher rates in comparison to existing baselines. While our method can show great benefits and potentials, there are open research questions yet to be addressed. First, we plan to evaluate how it can handle sim2real transfer in which a policy trained extensively in simulation can be evaluated in real bin-picking scenarios. Second, beside the two grasp and shift primitives we would like to see how more primitives, e.g. hand-over or throwing, can also be integrated and learnt in an end-to-end fashion using this framework.

\bibliographystyle{IEEEtran}
\bibliography{IEEEabrv,references}

\end{document}


\maketitle

\section{Network setting}
The specific network architecture details of the different modules are as follows:
\newline
\textbf{Pixel Encoder}: C(3,64)-MP-RB(128)-MP-RB(256)-RB(128)-UP-RB(64)-UP-C(64)
\newline
\textbf{Action Bridge}: C(64)-RELU-C(64)
\newline
\textbf{State Bridge}: C(64)-RELU-C(64)
\newline
\textbf{Actor}: C(256)-RELU-C(256)-RELU-C(256)-RELU-C($\left|a^{\phi}\right|$)
\newline
\textbf{Critic}: C(256)-RELU-C(256)-RELU-C(256)-RELU-C(1)
\newline
where C(k,c) denotes a 2D convolutional layer with k×k filters and c  channels, RB(c) denotes a resnet block with two 2D convolutional layers using 3x3 filters and c channels, MP denotes a 3×3 max-pooling layer  with stride 2, UP denotes a bi-linear upsampling layer using scale 2, and RELU is the Rectified Linear Unit activation.
\newline

\section{Training setting}

For both primitives, we use an Adam optimizer with a learning rate of $10^{-4}$ and a mini-batch size of 16. The coefficient $\alpha$ is  initialized with a value of 0.01 and optimized as described in \cite{DBLP:journals/corr/abs-1812-05905}, also using the Adam optimizer with  a learning rate of $10^{-4}$. We additionally control the exploration by using an $\epsilon$-greedy strategy, which balances between the use of the heuristic policy described in section 4.3 and the trained policy. The parameter $\epsilon$ starts at 0.9, and linearly scheduled to decrease to 0.2 within 2000 environment steps. 
During training we spawn a random number of $1-4$ objects in random poses and locations in the bin. We train on two different object types, cubes and rods. 
As described in the main text, we consider a short time horizon of $H=2$ for both primitives with a discount factor of $\gamma=0.99$.

\section{Additional results}
We evaluated the overall grasp performance in two modifications of the original experiments reported in experimental section of the manuscript. In one modified setting, we suppress the shifting primitive and allow only grasping actions, and in the other we use unseen objects.

\textbf{Grasp Only}
 In this setting, we wish to evaluate the contributions of the shifting primitives on the overall performance. For that purpose, we evaluate only grasp actions in each step using the same grasping network that was used in the original experiment. Specifically for our method, we use the network of the last step that predicts the grasping value only for the current step. Tables \ref{tab:results_random_grasp_only} and \ref{tab:results_challenging_grasp_only} summarize the performance metrics for random setting and the challenging settings respectively. The corresponding results of the original experiment that appears in parenthesis are repeated for reference.
\begin{table}[h]
    \centering
    \caption{Experimental Results for grasp-only ablation for the Random Setting (Mean $\%$)}
    \label{tab:results_random_grasp_only}
    \begin{tabular}{c|c|c|c|c}
        \toprule
        Method & Clearance & Completion  & Grasp Success  & Action Efficiency\\
         \midrule
        Berscheid et al. \cite{berscheid2019robot} & $68.3(72)$ &    $21.7(32)$   & $27.4(30)$ & $64.5(70)$
        \\
        Ours   & $85(86)$ & $51(60)$       & $55(55)$  & $85(80)$ 
        \\
        \bottomrule
    \end{tabular}
\end{table}

\begin{table}[h]
    \centering
    \caption{Experimental Results for the grasp-only ablation for the Challenging Setting (Mean $\%$)}
    \label{tab:results_challenging_grasp_only}
    \begin{tabular}{c|c|c|c|c}
        \toprule
        Method & Clearance &Completion  & Grasp Success  & Action Efficiency\\
         \midrule
        Berscheid et al. \cite{berscheid2019robot}  & $23.8(48)$ &    $1.6(1)$   & $7.7(15)$ & $62.3(66)$  
        \\
        Ours & $72(75)$ &    $30(39)$   & $33(37)$ & $72(68)$ 
        \\
        \bottomrule
    \end{tabular}
\end{table}

A comparison of these results with the original experiment underline the benefit resulting from an inclusion of the shifting primitive. For our method, we see improvements in both the random setting and even more pronounced in the challenging setting in the clearance and the completion metrics when shifting is enabled. Interestingly, the action efficiency is lowered by an inclusion of the shifting primitive. This counter-intuitive result can however be explained by the fact that only completed episodes, i.e. episodes in which the agent managed to empty the bin completely, contribute to this metric. In the grasp-only setting this is only possible in scenes, where all objects are graspable without shifting, whereas in the full setting also situations requiring a shift can contribute. In the latter, the action efficiency is naturally lower.
For the baseline we see the same improvement of clearance and completion. However we do not see the decrease in action efficiency, which, following the same line of argumentation as above, may indicate that the shifting primitive is not as successful in increasing the graspability of objects. 
Concerning the grasp success the situation is similar: while and inclusion of the shifting primitive leads to an increase in the grasp success for the baseline, it counter-intuitively shows no effect on the grasp success of our method. We can explain this result by the fact that our grasping primitive is already very dexterous and manages to grasp objects even in challenging situations. Therefore, shifts are necessary only quite rarely and averaged over all episodes the impact of successful shifts is hardly visible. 

\textbf{Unseen Objects}
Here, we wish to evaluate the generalization capability of our method. To this end, we apply the same shifting and grasping networks that were used in the original experiment on objects that did not appear during training. Specifically, in training we used cubes and rods, while in this evaluation setting we use cans, elongated boxes, balls, and tubes. We summarize the results of these experiments for both random and challenging setting in Tables \ref{tab:results_random_unseen} and \ref{tab:results_challenging_unseen} respectively. The corresponding results from the original experiment are again repeated in parenthesis for reference.
\begin{table}[h]
    \centering
    \caption{Experimental Results for unseen objects (orig. setting) for the Random Setting (Mean $\%$)}
    \label{tab:results_random_unseen}
    \begin{tabular}{c|c|c|c|c}
        \toprule
        Method & Clearance & Completion  & Grasp Success  & Action Efficiency\\
         \midrule
        Berscheid et al. \cite{berscheid2019robot} & $31.9(72)$ &    $1.5(32)$   & $43.5(30)$ & $41.9(70)$
        \\
        Ours   & $82.4 (86)$ & $53(60)$       & $46(55)$  & $73(80)$ 
        \\
        \bottomrule
    \end{tabular}
\end{table}

\begin{table}[h]
    \centering
    \caption{Experimental Results for unseen objects (orig. setting) for the Challenging Setting (Mean $\%$)}
    \label{tab:results_challenging_unseen}
    \begin{tabular}{c|c|c|c|c}
        \toprule
        Method & Clearance & Completion  & Grasp Success  & Action Efficiency\\
         \midrule
        Berscheid et al. \cite{berscheid2019robot} & $33.7(48)$ &    $0.9(1)$   & $11.6(15)$ & $80(66)$
        \\
        Ours   & $77(75)$ & $49(39)$       & $39(37)$  & $62(68)$ 
        \\
        \bottomrule
    \end{tabular}
\end{table}

As could be expected, most of the performance metrics decreased in light of the unseen objects for both methods. However, it is apparent that while the performance degradation is quite significant for the baseline (particularly in the random setting), our method exhibits only minor reductions in performance. Our observations reveal that these results are only mildly related to the 6D primitives inference in a direct manner, and can only speculate that the better generalization of our method has more to do with the networks architecture. For the challenging setting, it appears that there is even a slight improvement of performance using our method with the unseen objects, which is counter-intuitive. However, we attribute this finding to the particular objects whose shape pose smaller challenge even in proximity to the wall. 

\small
\bibliography{references}